\pdfoutput=1
\documentclass{article}

\usepackage[dblblindworkshop,final]{neurips_2026}
\workshoptitle{Resource-Aware Agentic AI}

\usepackage[utf8]{inputenc}
\usepackage[T1]{fontenc}
\usepackage{hyperref}
\hypersetup{pdftitle={Batching the Reflective Optimization Loop: Parallel Proposals Make GEPA Faster and Better},pdfauthor={Jialin Zhang, Lakshya A Agrawal, Shangyin Tan, Donghyun Lee, Lutfi Eren Erdogan, Dan Klein, Koushik Sen, Alex Dimakis, Matei Zaharia}}
\usepackage{url}
\usepackage{booktabs}
\usepackage{amsfonts}
\usepackage{amsmath}
\usepackage{nicefrac}
\usepackage{microtype}
\usepackage{graphicx}
\usepackage{xcolor}

\graphicspath{{figures/}}

\newcommand{\pxn}{P{\times}N}

\title{Batching the Reflective Optimization Loop:\\Parallel Proposals Make GEPA Faster and Better}

\author{%
  Jialin Zhang\thanks{Equal contribution.}%
  \;\; Lakshya A Agrawal\footnotemark[1]%
  \;\; Shangyin Tan\footnotemark[1]%
  \;\; Donghyun Lee\footnotemark[1]\\
  \textbf{Lutfi Eren Erdogan} \;\; \textbf{Dan Klein} \;\; \textbf{Koushik Sen}
  \;\; \textbf{Alex Dimakis} \;\; \textbf{Matei Zaharia}\\
  University of California, Berkeley\\
  \texttt{benzhang04@berkeley.edu}\\
}

\begin{document}

\maketitle

\begin{abstract}
Reflective text optimization makes agentic systems better, but the
optimization loop is itself a resource-hungry workload whose sequential
structure leaves parallel evaluation capacity idle.
We extend the open-source GEPA optimizer to propose and evaluate
a batch of candidates on each
optimization step instead of one candidate at a time. In our sweep on two
tasks, most batched runs finished in half the wall-clock time or less, and the
fastest in about a quarter to a third. Batched settings also achieved higher
held-out test scores, from 68.9\% to 72.1\% on LiveBench-Math (with $2{\times}2$)
and from 49.0\% to 60.0\% on HoVer (with $8{\times}1$).
A simple runtime model explains the measured speedups under a finite
worker pool, and batched settings also spend their metric-call and dollar
budgets more efficiently. We frame the axes along which a reflective
optimization step can scale in parallel, a first look at allocating a fixed
optimization budget across them.
\end{abstract}

\begin{figure}[h]
  \centering
  \includegraphics[width=\linewidth]{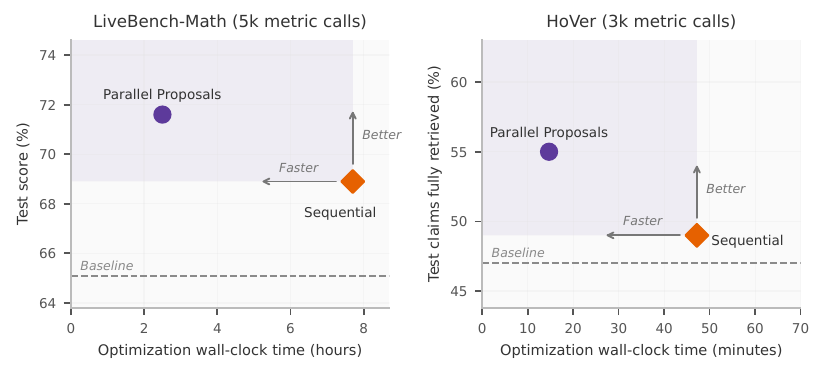}
  \caption{At the same metric-call budget, parallel proposals are faster and
  better than single mutation, achieving up to about a 3 to 4$\times$ speedup,
  while scoring up to 11 points higher on the held-out test set.}
  \label{fig:teaser}
\end{figure}

\section{Introduction}

Optimizing an agentic AI system is itself a resource problem. Reflective
optimizers such as GEPA \citep{agrawal2025gepa} improve an agent by proposing
textual mutations of its prompts or code, scoring them with rollouts, and
keeping the winners, so a run consumes LLM calls, dollars, and, above all,
wall-clock time. GEPA is sample-efficient in the number of rollouts it needs,
but its loop is sequential.
Running GEPA on a task can take hours because each optimization step waits for
a proposal and its evaluation before the next step begins. In each iteration,
GEPA samples a parent, proposes a mutation, evaluates it on a mini-batch, and,
if it improves on its parent, evaluates it on the full validation set.

We extend GEPA with batched parallel proposals. Instead of advancing one
proposal at a time, a step can propose several candidates and dispatch their
evaluations concurrently. This allows for evaluating more proposals at a time,
so the same budget of metric calls finishes in less wall-clock time, and it
delivers better results, by exploring more of the search tree per step. In our
experiments, batching cut the wall-clock time by roughly 3 to 4$\times$, while
raising held-out test scores from 68.9\% to 72.1\% on LiveBench-Math and from
49.0\% to 60.0\% on HoVer (Figure~\ref{fig:teaser}).

Faster and better optimization makes GEPA more accessible and practical for
more tasks, and lets users spend more budget in the same amount of time. Beyond
the batching mechanism itself, we also introduce the axes of parallel
scaling in reflective optimization, a potential area for future study
(Appendix~\ref{app:axes}).

\section{How parallel proposals work}
\label{sec:method}

\begin{figure}[t]
  \centering
  \includegraphics[width=\linewidth]{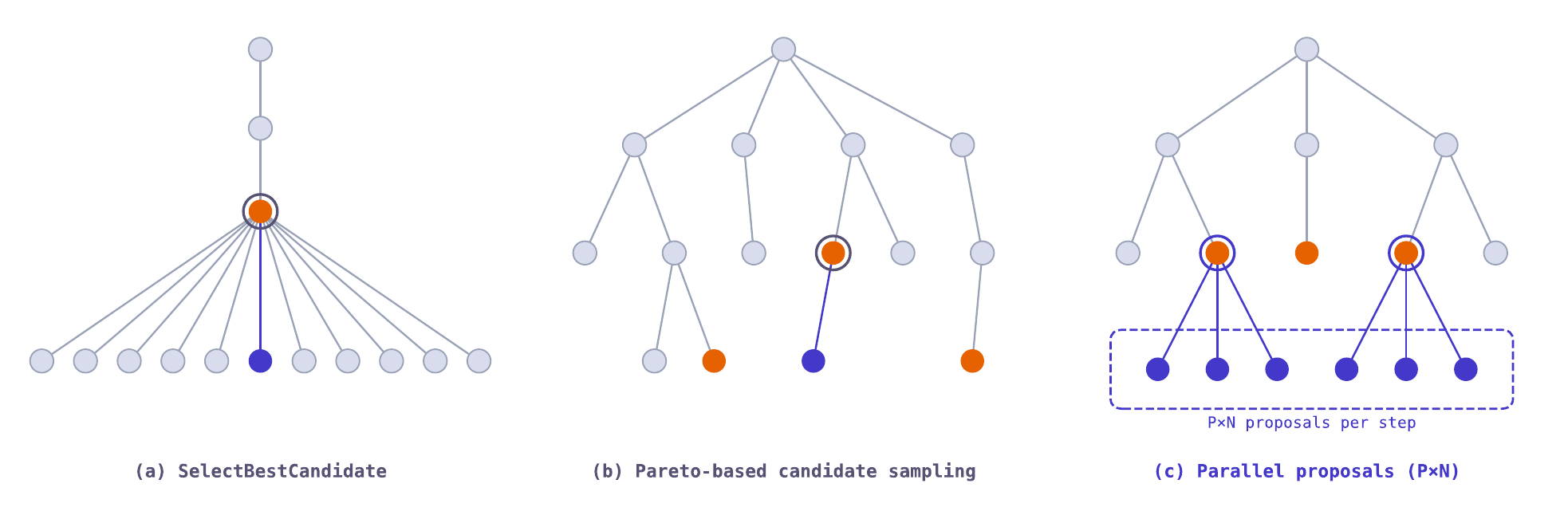}
  \caption{From best-candidate selection to parallel proposals.
  SelectBestCandidate extends only the current best candidate, GEPA's
  Pareto-based candidate sampling explores a broader tree one proposal at a
  time, and parallel proposals ($\pxn$) extend several frontier candidates at
  once.}
  \label{fig:concept}
\end{figure}

The mechanism is closely analogous to batch Bayesian optimization
\citep{ginsbourger2010kriging}, which proposes and evaluates a batch of
candidates per round rather than adapting after every single one. Similar to
how GEPA uses a Pareto frontier instead of a single best numerical score to
select candidates, parallel proposals push the idea further by drawing several
extensions of the frontier within each step
(Figure~\ref{fig:concept}), thus reducing how often the search adapts to the
validation set. Prior work \citep{dwork2015reusable} also shows that repeatedly
steering decisions with one fixed holdout can inflate its apparent performance,
so committing to more proposals simultaneously should transfer better beyond
the validation set.

\begin{figure}[t]
  \centering
  \includegraphics[width=\linewidth]{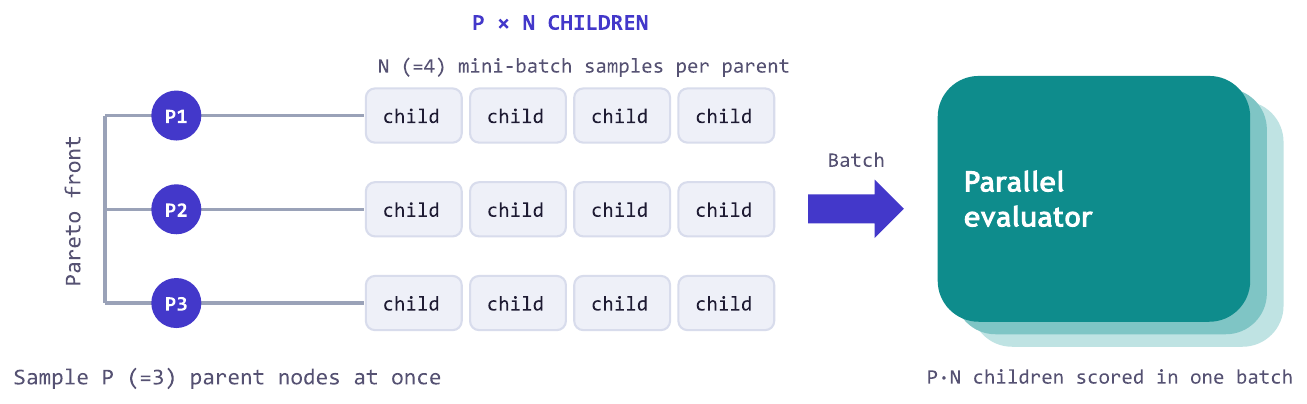}
  \caption{One batched iteration. GEPA samples $P$ parents from the frontier,
  draws $N$ reflective mutations for each parent, and scores all $P \cdot N$
  children in one parallel evaluation. This lets GEPA propose more candidates
  in each iteration while paying the iteration latency once.}
  \label{fig:pxn}
\end{figure}

Standard GEPA advances one proposal per step, extending a single candidate
drawn from its Pareto frontier. Parallel proposals sample several parents from
the frontier, draw several reflective mutations of each, and score all of the
proposals concurrently (Figure~\ref{fig:pxn}).

In our $\pxn$ sampling strategy, one GEPA step
\begin{enumerate}
  \item samples $P$ parents from the current Pareto frontier;
  \item draws $N$ mini-batches per parent and evaluates each parent on them
        through \texttt{batch\_evaluate()}, producing $P \cdot N$ reflection
        requests;
  \item dispatches the reflection requests concurrently;
  \item screens the proposals by batch-evaluating them again on their
        mini-batches;
  \item evaluates accepted candidates on the full validation set in parallel,
        then updates the frontier.
\end{enumerate}

\section{Runtime analysis}
\label{sec:runtime}

Write $T(k)$ for the wall-clock time of a run with $k = P \cdot N$ proposals
per step, so that $T(1)$ is a single-mutation run. It is the sum of the time
spent on each iteration. Every iteration incurs a step latency
$L_{\text{step}}$ for the $k$ reflection calls and $k$ mini-batch evaluations,
which run concurrently. Whenever one or more proposals beat their parent on the
mini-batch, the iteration additionally pays a full-validation latency
$L_{\text{val}}$ to evaluate all accepted proposals on the validation set in
parallel.

The metric-call budget determines the total number of proposals a run can
afford, whether parallel or sequential (if we assume the candidate acceptance
rate stays the same). The number of iterations is the number of proposals
divided by $k$, so a width-$k$ run needs about $1/k$ as many iterations as
single mutation. The main bottleneck is full validation. If each proposal is
accepted with probability $a$, and proposal outcomes are independent, then an
iteration triggers full validation with probability
\begin{equation}
  q_k = 1 - (1-a)^k,
\end{equation}
which grows with $k$. Multiplying the per-iteration cost by the number of
iterations gives the speedup of a width-$k$ run over single mutation,
\begin{equation}
  \frac{T(1)}{T(k)} \approx
  \frac{k\,(L_{\text{step}} + a\,L_{\text{val}})}{L_{\text{step}} + q_k\,L_{\text{val}}}.
  \label{eq:speedup}
\end{equation}

In practice, two effects slow wide steps down. First, the reflection stage
takes as long as the slowest of its $k$ concurrent calls, so $L_{\text{step}}$
grows with width. Second, evaluation can be limited by the worker pool. A
validation stage can carry up to $k$ accepted candidates, each evaluated on all
$V$ validation examples, so with $W$ concurrent workers and a per-rollout
latency of $T_e$,
\begin{equation}
  L_{\text{val}} \approx
  \begin{cases}
    T_e & \text{if } kV \le W, \\
    (kV/W)\,T_e & \text{if } kV > W.
  \end{cases}
  \label{eq:workers}
\end{equation}
The predicted curve in Figure~\ref{fig:scaling} uses the expected number of
accepted candidates, $a k$, in place of the worst case $k$, so its validation
term per iteration is $\max\!\big(q_k\,L_{\text{val}},\; a k V T_e / W\big)$
with $L_{\text{val}}$ evaluated at width one, and it holds $L_{\text{step}}$
fixed at its measured single-mutation value.

According to strong scaling \citep{amdahl1967validity}, a run with a budget of
$B$ metric calls needs at least $B \cdot T_e / W$ of wall-clock time for
evaluation alone when $W$ is fixed. This bound is most relevant for
system-bound workloads, where available CPU or cluster capacity is genuinely
fixed. In agent optimization workloads evaluated against large-scale inference
services, however, the worker pool can often be increased substantially. By
raising the number of workers in the pool, we expect to further lower this
bound.

\section{Results}
\label{sec:results}

We evaluated parallel proposals on LiveBench-Math \citep{white2025livebench}
and HoVer \citep{jiang2020hover}, where the optimized agent is selected using a
validation set and then measured on a held-out test set. For both tasks, we
used \texttt{gpt-5-mini} as the proposer. As in standard GEPA runs, we measured
the optimization budget by the number of metric calls. One metric call
corresponds to evaluating one candidate on one example. Every setting on a task
received the same total metric-call budget.
LiveBench-Math optimizes a single solver prompt over competition math
problems with a budget of 5,000 metric calls, and HoVer optimizes the two
components of a four-hop retrieval agent over a multi-hop claim verification
corpus with a budget of 3,000 metric calls. Appendix~\ref{app:tasks} gives the
full task setups, splits, and metrics.

\subsection{Batching cuts the wall-clock time and scores higher on held-out tests}

\begin{figure}[t]
  \centering
  \includegraphics[width=\linewidth]{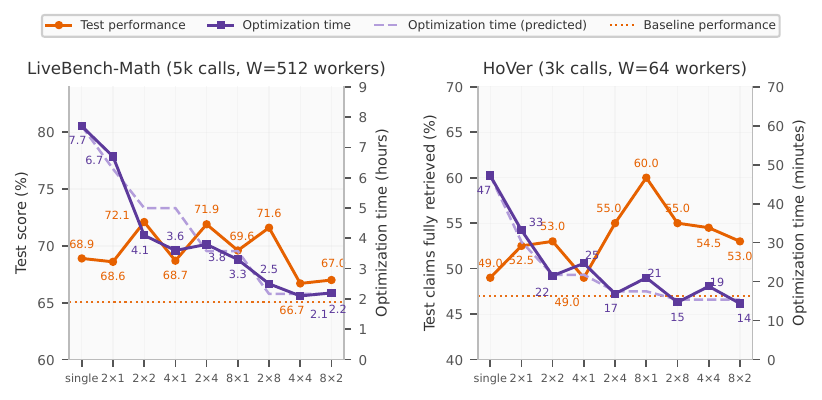}
  \caption{As the per-step width $P \cdot N$ scales up, runtime falls with
  diminishing returns. The finite-worker prediction matches the measured
  runtimes. Most settings perform as well as or better than single mutation on
  test, and the best setting scores much higher.}
  \label{fig:scaling}
\end{figure}

We ran evaluations with up to 512 concurrent workers on LiveBench-Math and 64
on HoVer. The measured runs (Figure~\ref{fig:scaling}) align with the runtime
model above, cutting the wall-clock time by about 3 to 4$\times$ at 16-way
parallelism. For example, on LiveBench-Math, moving from single mutation to
$2{\times}2$ cut the number of iterations by 4.9$\times$ (219 to 45), but the
fraction of iterations that triggered full validation rose from 17\% to 53\%,
so the run gained a 1.9$\times$ speedup (7.7 to 4.1 hours).

Figure~\ref{fig:scaling} also shows that most batched settings match or beat
single mutation on the held-out tests, and the best settings reach 72.1\%
against 68.9\% on LiveBench-Math ($2{\times}2$) and 60.0\% against 49.0\% on
HoVer ($8{\times}1$).

\subsection{Scaling parents ($P$) vs.\ mutations ($N$)}

In principle, larger $P$ extends more members of the frontier at once, which
should help when no single generally good candidate exists and the frontier
holds genuinely different specialists worth advancing in parallel, and larger
$N$ draws more mutations with different mini-batches, which should help when
the dataset is rich enough to expose many distinct directions to improve one
candidate. LiveBench-Math is the second case, with problems spanning multiple
areas, so giving each parent several mutations (larger $N$) transferred better
to test than spreading single mutations across more parents (larger $P$). HoVer
additionally keeps a more complementary candidate pool, where different
candidates succeed on different claims, and its test scores tend to grow with
both $P$ and $N$. With $P$ held at 2, test scores rise with $N$ from 68.6 at
$2{\times}1$ to 71.6--72.1 at $2{\times}2$ through $2{\times}8$ on
LiveBench-Math, and from 52.5 to 55.0 on HoVer
(as shown in Figure~\ref{fig:scaling}). Generally, we observe scaling $N$
improves performance on both tasks, even though the highest-scoring split
remains task-dependent ($2{\times}2$ on LiveBench-Math, $8{\times}1$ on HoVer).
We leave further ablations on other budgets and parallelisms for future work.

\subsection{Batched settings are more budget-efficient}

We also probed how efficiently each setting spends its budget during the run.
Here we compare a batched setting, $2{\times}4$, against single mutation on
both tasks.

\begin{figure}[t]
  \centering
  \includegraphics[width=\linewidth]{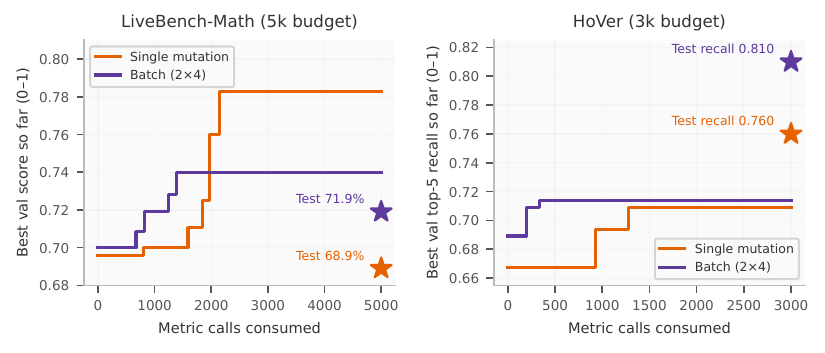}
  \caption{Best validation quality against metric calls consumed, for single
  mutation and $2{\times}4$. $2{\times}4$ dominates small budgets and scores
  higher on the held-out test set.}
  \label{fig:budget}
\end{figure}

\paragraph{More quality per metric call.}
$2{\times}4$ dominates small budgets. On LiveBench-Math it reaches validation
scores that single mutation needs about 2,000 calls to match, and on HoVer it
leads at every budget (Figure~\ref{fig:budget}). Single mutation raises
validation scores later in the budget, but still scores lower on the held-out
test set on both tasks.

On the held-out test sets, $2{\times}4$ won 3.0pp over single mutation on
LiveBench-Math and 6.0pp on HoVer. The validation scores tell a different
story. On LiveBench-Math, single mutation scored higher on validation but lower
on test. Its generalization gap (the drop from validation score to test score)
was nine points, against only two for $2{\times}4$. On HoVer, both settings'
test scores landed above their validation scores, and $2{\times}4$ ended higher
on both. This is consistent with the hypothesis above, that committing to more
proposals before adapting to the validation set transfers better beyond it.

\begin{figure}[t]
  \centering
  \includegraphics[width=\linewidth]{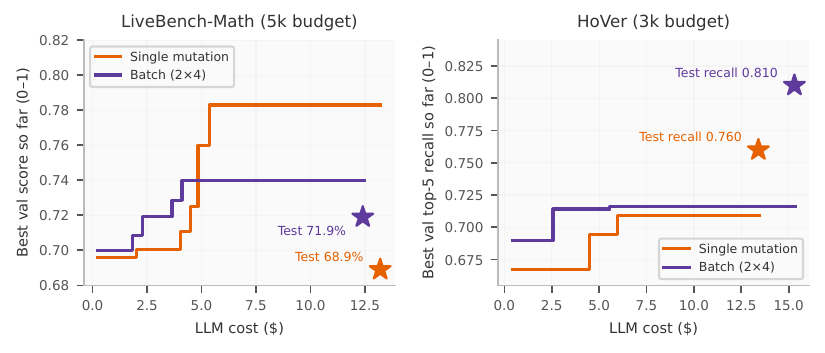}
  \caption{Best validation quality against total LLM spend, solver and
  reflection calls combined. The batched setting reaches strong validation
  quality within the first few dollars.}
  \label{fig:cost}
\end{figure}

\paragraph{More quality per dollar.}
By measuring the total LLM spend during optimization (sum of solver calls and
reflection calls), we found that the batched setting is also cost-efficient,
reaching strong validation scores within the first few dollars
(Figure~\ref{fig:cost}). On LiveBench-Math, $2{\times}4$ reaches its final
validation score within about \$4 of spend, and on HoVer it passes single
mutation's final validation score within about \$2.6. The total spend stays
comparable across settings, within the \$12 to \$16 range.

\section{Related work}
\label{sec:related}

Compile-time optimizers for LLM programs, including the DSPy family
\citep{khattab2024dspy, opsahlong2024optimizing} and reflective evolution in
GEPA \citep{agrawal2025gepa}, treat prompts and program text as the parameters
of an agentic system; evolutionary search with LLM mutation operators likewise
drives systems such as AlphaEvolve \citep{novikov2025alphaevolve}. Our batching
strategy adapts a classical idea from batch Bayesian optimization
\citep{ginsbourger2010kriging}, where the optimizer commits to several
evaluations per round instead of adapting after each one. The generalization
behavior we observe connects to adaptive data analysis, where limiting the
number of adaptive rounds against a fixed holdout preserves validity
\citep{dwork2015reusable, blum2015ladder}. Finally, allocating a fixed budget
across the width and depth of an optimization step echoes work on scaling
test-time compute \citep{snell2024scaling}, applied here to the optimization
loop rather than to a single query.

\section{Conclusion}

Parallel proposals batch GEPA's reflective optimization loop. Across
LiveBench-Math and HoVer, batched settings cut optimization wall-clock time by
about 3 to 4$\times$, most of them matched or beat single mutation on held-out
tests, and a batched setting reached strong validation quality with fewer
metric calls and fewer dollars. A
simple runtime model predicts the measured speedups and shows where the
remaining time goes. The axes of parallel scaling (Appendix~\ref{app:axes})
frame the open question of how to allocate a fixed optimization budget across
parents, mutations, reflection mini-batches, and reflection compute. Our
implementation builds on the open-source \texttt{gepa} package as a plug-in
sampling strategy (Appendix~\ref{app:usage}).

\bibliographystyle{plainnat}
\bibliography{references}

\appendix

\section{Axes of parallel scaling}
\label{app:axes}

\begin{figure}[h]
  \centering
  \includegraphics[width=\linewidth]{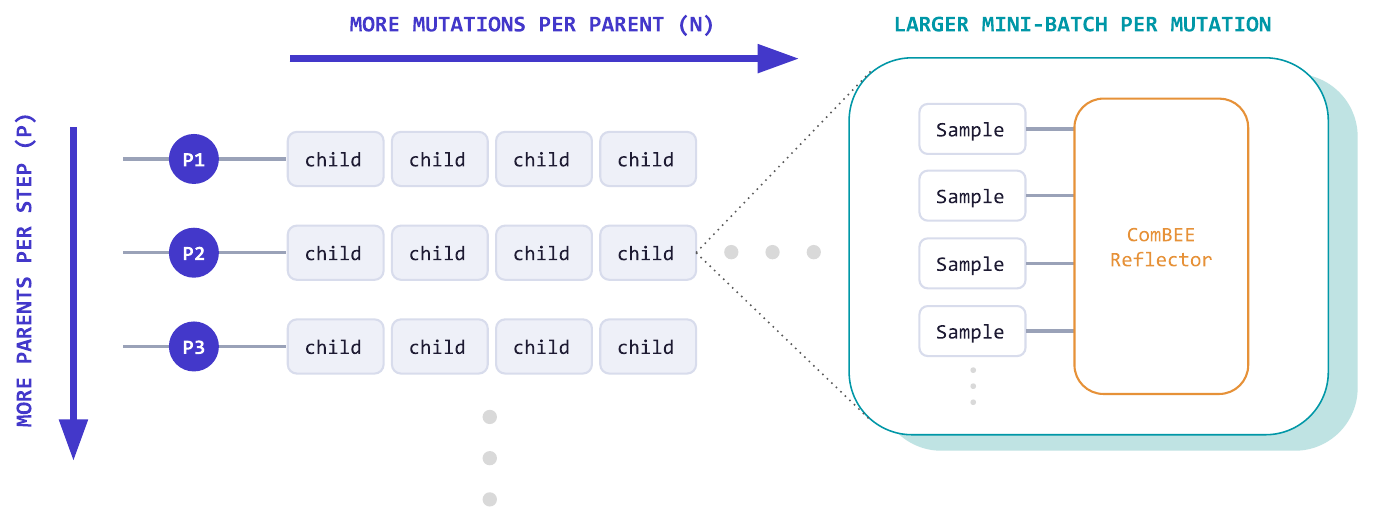}
  \caption{Scaling axes for the optimization step: more parents per step ($P$),
  more mutations per parent ($N$), and a larger reflection mini-batch per
  mutation (ComBEE).}
  \label{fig:axes}
\end{figure}

Parallel proposals fit into the bigger picture of scaling each reflection
optimization step along several axes (Figure~\ref{fig:axes}). $P$ controls how many programs are
selected for mutation at each step, and $N$ controls how many mutations are
proposed for each selected program. A third axis is the size of the reflection
mini-batch per mutation. Approaches like ComBEE \citep{li2026combee} scale
along this axis by reflecting on larger mini-batches of examples and rollouts
through a map-reduce-style pipeline. Finally, a fourth axis is the compute
spent within each reflection step. It can be turned up by using stronger
reflection programs ranging from a single LM call to chain-of-thought,
multi-step pipelines, or agentic workflows such as Claude
Code.\footnote{\url{https://gepa-ai.github.io/gepa/guides/claude-cli-as-proposer/}}
In each case, more compute is spent to produce a higher-information mutation at
each step.

With this, GEPA offers four independently tunable dials for spending more
optimization compute per step, namely $P$, $N$, reflection mini-batch (ComBEE),
and reflection compute (reflection LM, agentic proposers). They interact
through a shared optimization budget (metric calls, tokens, or dollars).
Parallel proposals are one way to spend it. The open question is how to
allocate a fixed optimization budget across these dials, and this work is a
first look at that trade-off.

\section{Task setup details}
\label{app:tasks}

\paragraph{LiveBench-Math.}
LiveBench-Math \citep{white2025livebench} asks a model to solve competition
math problems (AMC and AIME questions, symbolic algebra, and olympiad
problems), graded by LiveBench's own scorers, with the
Terrarium\footnote{\url{https://github.com/gepa-ai/terrarium}} split of 100
training, 100 validation, and 168 test problems. Budget: 5,000 metric calls,
each one a \texttt{gpt-4.1-mini} solution attempt.

\paragraph{HoVer.}
HoVer \citep{jiang2020hover} asks a system to gather the Wikipedia pages needed
to verify a multi-hop claim. We optimize the two components (a query writer and
a note taker) of a four-hop \texttt{gpt-4.1-mini} retrieval agent over a BM25
index of 5.2 million 2017 Wikipedia abstracts, on three-hop claims split into
200 training, 150 validation, and 200 test claims; one rollout makes about
eight calls. During optimization, GEPA scores each rollout by the fraction of
the claim's three gold pages that appear in the retrieved pages (top-5 recall),
as shown by the validation curves and test stars in Figures~\ref{fig:budget}
and~\ref{fig:cost}. The headline numbers in the text and in
Figures~\ref{fig:teaser} and~\ref{fig:scaling} are the share of test claims
with all three gold pages retrieved. Budget: 3,000 metric calls, each one a
full program rollout.

\section{Using parallel proposals}
\label{app:usage}

Our implementation plugs into the sampling-strategy interface of the
open-source \texttt{gepa} engine and is enabled with the setting change below.
The sampling strategy says how many candidates to propose per step, and the
selection strategy says which of the improved candidates to keep. For example, two parents with two mutations each
gives four candidates per step.

\begin{small}
\begin{verbatim}
from gepa.optimize_anything import optimize_anything, OptimizeAnythingConfig
from gepa.strategies.proposal_sampling import PxNSampling
from gepa.strategies.proposal_selection import AllImprovements

config = OptimizeAnythingConfig(
    engine="gepa",
    max_concurrency=64,
    engine_config={
        "engine": {
            "sampling_strategy": PxNSampling(p=2, n=2),
            "selection_strategy": AllImprovements(),
        },
        "reflection": {"reflection_lm": "gpt-5-mini"},
    },
)

result = optimize_anything(
    seed_candidate=seed, evaluator=evaluate,
    dataset=trainset, valset=valset, objective=objective, config=config,
)
\end{verbatim}
\end{small}

GEPA by default calls the user-supplied \texttt{evaluate} function in parallel, so
\texttt{max\_concurrency} should be set high enough to match the capacity of the evaluator
and inference provider. For system-bound evaluations,
\texttt{max\_concurrency} may be constrained by available CPU, GPU, or cluster
capacity. For API-backed optimization, much larger worker pools may be
practical, subject to provider rate limits and budget. Optionally, users may
provide a custom \texttt{batch\_evaluate} function via the
\texttt{batch\_evaluator} argument. Since the GEPA engine remains
single-threaded and dependency-free, the choice of parallelism is entirely up to
the user. Any concurrency or distributed-execution framework
(Harbor, Ray, Slurm, Modal, Daytona, and more) can run the evaluations, and any
inference backend that supports batch completion to run reflections in
parallel, with no changes to GEPA itself. Users may also choose or define other
sampling and selection strategies.

\end{document}